\documentclass[letterpaper, 10 pt, conference]{ieeeconf}  

\usepackage{cite}
\usepackage{graphicx}
\usepackage{textcomp}

\usepackage{amsmath,amssymb,amsfonts}

\usepackage{amsthm}

\usepackage{algorithm}
\usepackage{algpseudocode}
\algrenewcommand\algorithmicrequire{\textbf{Input:}}
\algrenewcommand\algorithmicensure{\textbf{Output:}}

\usepackage{booktabs}
\usepackage{multirow}
\usepackage[table]{xcolor} 
\usepackage{makecell}

\newtheorem*{remark}{Remark}

\IEEEoverridecommandlockouts                              

\title{\LARGE \bf
Boosting Action-Information via a Variational Bottleneck on Unlabelled Robot Videos
}

\author{Haoyu Zhang and Long Cheng
\thanks{The authors are all with the School of Artificial Intelligence, 
University of Chinese Academy of Sciences, Beijing 100049, China. They are also 
with the State Key Laboratory of Multimodal Artificial Intelligence Systems, 
Institute of Automation, Chinese Academy of Sciences, Beijing 100190, China. All 
correspondence should be addressed to the corresponding author, Dr. Long Cheng 
(long.cheng@ia.ac.cn).}
}

\begin{document}

\maketitle
\thispagestyle{empty}
\pagestyle{empty}

\begin{abstract}

Learning from demonstrations (LfD) typically relies on large amounts of action‐labeled expert trajectories, which fundamentally constrains the scale of available training data. A promising alternative is to learn directly from unlabeled video demonstrations. However, we find that existing methods tend to encode latent actions that share  little mutual information with the true robot actions, leading to suboptimal control performance. To address this limitation, we introduce a novel framework that explicitly maximizes the mutual information between latent actions and true actions, even in the absence of action labels. Our method leverage the variational information-bottleneck to extract action-relevant representations while discarding task-irrelevant information. We provide a theoretical analysis showing that our objective indeed maximizes the mutual information between latent and true actions. Finally, we validate our approach through extensive experiments: first in simulated robotic environments and then on real‐world robotic platforms, the experimental results demonstrate that our method significantly enhances mutual information and consistently improves policy performance.

\end{abstract}

\section{INTRODUCTION}

Training from diverse real-world robot datasets has proven effective for learning robust control policies across a variety of tasks \cite{sec1_b1}. However, collecting such datasets typically requires extensive human teleoperation or specialized instrumentation, which limits both the scale and the diversity of available demonstrations. By contrast, internet-scale video repositories contain abundant, unlabeled examples of human behaviors and physical interactions, offering a promising alternative for scaling learning-from-demonstration (LfD) methods.

Most existing works on video-based LfD \cite{sec2_b10, sec2_b11, sec2_b12, sec2_b13, sec2_b14} adopt a variational autoencoder (VAE) paradigm: an inverse dynamics model (IDM) infers a latent action representation from consecutive observations, and a forward dynamics model predicts future observations conditioned on that latent. As we show in Sec. \ref{sec1_prob_stat}, however, the latent actions produced by this VAE framework capture only a small fraction of the true control commands’ information, which in turn degrades the performance of policies learned on these latents.

\subsection{Problem Statement}\label{sec1_prob_stat}
We perform a {\it quantitative evaluation} of LAPA’s \cite{sec2_b14} latent actions on eight MetaWorld \cite{sec1_metaworld} manipulation tasks (detailed in Sec. \ref{sec:experiment_numerical}).  Specifically, we fine-tune a publicly released LAPA model (pre-trained on OpenX video data) on MetaWorld frame pairs, then extract latent actions $z \in \mathcal{Z}$ for each frame transition and compare them to the ground-truth robot commands $a \in \mathcal{A}$. To quantify the alignment between each latent dimension $z_i$ and each true action channel $a_j$, we compute two metrics:


\begin{enumerate}
  \item {\bf Action entropy.}  We discretize each action channel \(a_j\) into \(n=256\) histogram bins, count frequencies \(\{c_k\}_{k=1}^n\), and form the empirical distribution \(p_k = c_k / N\).  The entropy is then
  \[
    H(a_j) \;=\; -\sum_{k=1}^n p_k \,\ln p_k\,.
  \]
  
  \item {\bf Mutual information.}  Likewise, each latent dimension \(Z_i\) is discretized into 256 bins.  We build the joint histogram \(c_{uv}\) over bins \(u\) of \(z_i\) and \(v\) of \(a_j\), normalize to \(p_{uv}=c_{uv}/N\), and compute
  \[
    I(z_i;a_j)
    \;=\;
    \sum_{u=1}^n\sum_{v=1}^n
      p_{uv}\,\ln\frac{p_{uv}}{p_{u}\,p_{v}}
    \quad
  \]
where $p_{u}=\sum_{v}p_{uv}, p_{v}=\sum_{u}p_{uv}.$ We then report the {\it capture ratio}
  \(\max_i I(z_i;a_j) / H(a_j)\).

  \item {\bf Pearson correlation.}  We compute the standard Pearson coefficient
  \[
    r(z_i,a_j)
    \;=\;
    \frac{\operatorname{Cov}\bigl(z_i,a_j\bigr)}
         {\sigma_{z_i}\,\sigma_{a_j}},
  \]
  using the empirical covariance over all samples.  For each channel \(j\) we report \(\max_i |r(z_i,a_j)|\).
\end{enumerate}

\begin{figure}[t]
  \centering
  \includegraphics[width=1.0\linewidth]{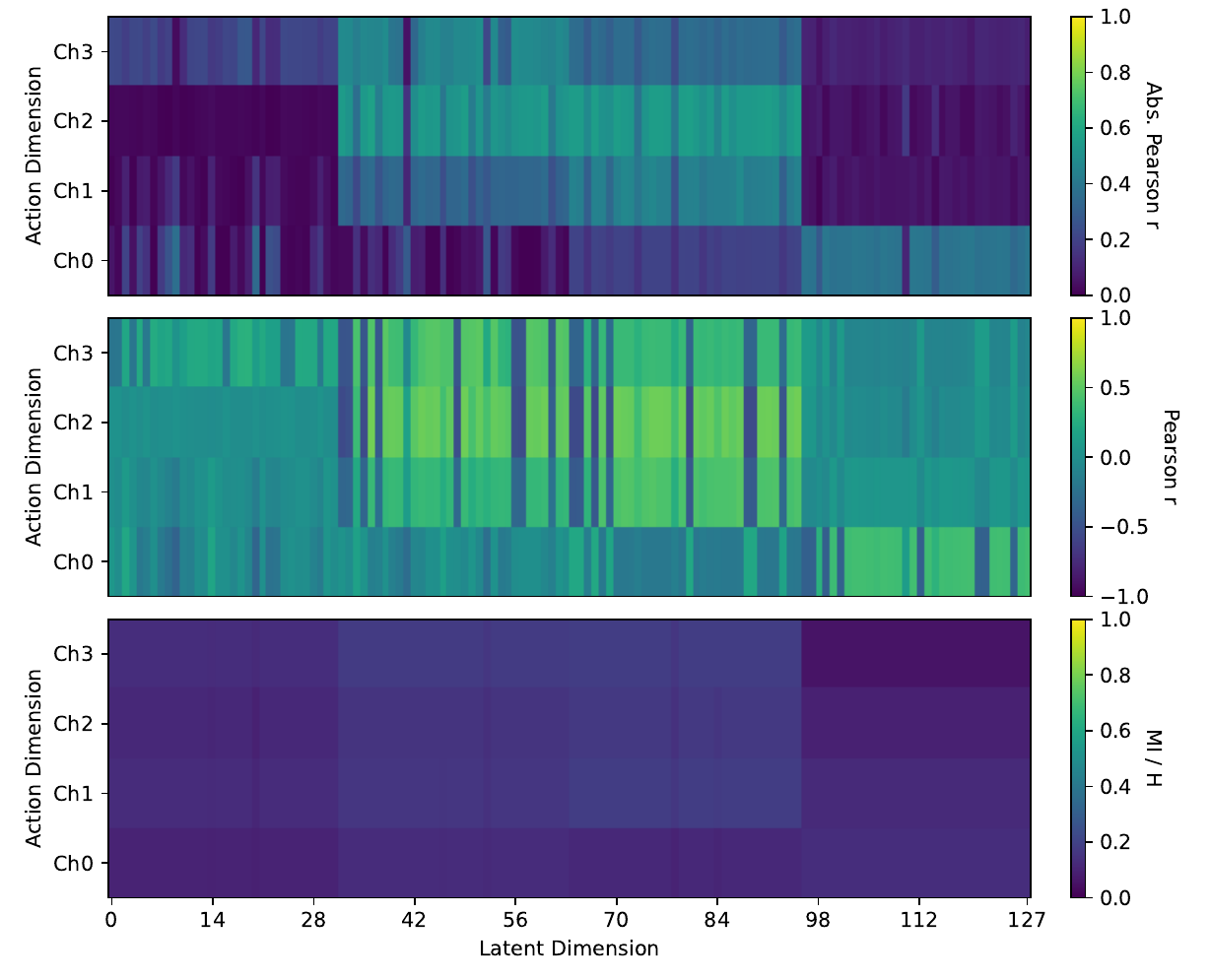}
  \caption{Heatmaps comparing LAPA’s latent dimensions and true actions on MetaWorld.  {\bf Top:} Absolute Pearson correlation $|r(z_i,a_j)|$.  {\bf Bottom:} Normalized mutual information capture ratio $I(z_i;a_j)/H(a_j)$.}
  \label{fig:comparison}
\end{figure}

Our results show that, although certain latent dimensions attain moderate Pearson correlation ($r\approx 0.4\!-\!0.57$), they only capture $13\%\!-\!18\%$ of the action entropy (see Table \ref{tab:comparison} and Figure \ref{fig:comparison}).  This low information capture explains why directly regressing policies from latent actions, as done in LAPO \cite{sec2_b11}, yields poor performance.

Based on this analysis, we address two key questions:
\begin{enumerate}
  \item \textbf{Unsupervised information enhancement:} How can we increase the mutual information between latent actions $\mathcal{Z}$ and true control commands $\mathcal{A}$ using only unlabelled video?
  \item \textbf{Few-shot policy learning:} Given a small set of action-labelled demonstrations, how can we best leverage the enriched latent actions to learn high-quality control policies?
\end{enumerate}

In this work, we present a framework that increases the mutual information between latent actions inferred from unlabeled videos and the corresponding ground-truth actions, thereby improving robot policy performance. Our contributions are:
\begin{enumerate}
\item We replace the VAE with a variational information bottleneck (VIB) formulation to infer latent actions, explicitly encouraging higher mutual information with the true actions.
\item We provide a theoretical analysis that proves the effectiveness of the proposed method by establishing conditions under which the formulation increases mutual information.
\item We conduct extensive experiments showing that the VIB framework yields higher mutual information and achieves superior performance across a range of robotic tasks.
\end{enumerate}

\begin{table}[t]\scriptsize
  \centering
  \caption{LAPA latent–action alignment on MetaWorld (8 tasks).}
  \label{tab:comparison}
  \begin{tabular}{lcccc}
    \toprule
     & \multicolumn{4}{c}{Dimension} \\
    Metric      & dim~0  & dim~1  & dim~2  & dim~3  \\
    \midrule
    $\max_i\lvert r(z_i,a_j)\rvert$ & 0.400 & 0.481 & 0.571 & 0.500 \\
    $\max_i I(z_i;a_j)/H(a_j)$~(\%) & 13.2\% & 18.2\% & 16.6\% & 18.4\% \\
    \bottomrule
  \end{tabular}
\end{table}

\begin{figure*}[!t] 
  \centering
  \includegraphics[width=1.0\textwidth]{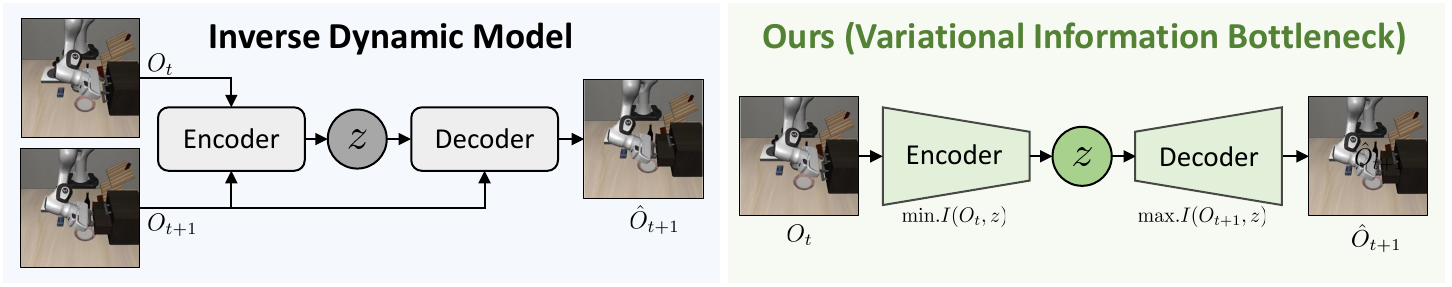}
  \caption{%
    \textbf{IDM vs. VIB.} 
    IDM encodes $\{x_t,x_{t+1}\}$ to $z$ and reconstructs $x_{t+1}$ from $(x_t,z)$. 
    Ours encodes $x_t$ to $z$ and reconstructs $x_{t+1}$ from $z$, enforcing an information bottleneck: minimize $I(z;x_t)$ and maximize $I(z;x_{t+1})$.
  }
  \label{fig:metaworld-result}
\end{figure*}

\section{RELATED WORKS}
\subsection{Learning from Videos}
Humans often learn by watching raw videos rather than from paired observation-action data, indicating that videos contain rich information about how to perform specific tasks. In contrast, robot learning typically relies on imitation learning, which requires precise action labels that are rarely available at scale. To address this limitation, prior studies have used egocentric human videos to pretrain visual representations for visuomotor control \cite{sec2_b1, sec2_b2, sec2_b3}; aligned or translated observation sequences \cite{sec2_b4, sec2_b5, sec2_b6}; performed online rollouts or used standard computer vision tools to predict future observations and derive control signals from videos \cite{sec2_b7, sec2_b8}; and explored video generative models pretrained on human videos for downstream robotic tasks \cite{sec2_b9}. Despite these advances, most methods learn visual priors rather than explicit control signals.

\subsection{Latent Action Models}
Prior work on latent action modeling typically trains IDMs to infer actions or latent surrogates from pairs of current and future observations. VPT \cite{sec2_b10} begins with a small labeled seed set to supervise an IDM and then scales to large unlabeled corpora via pseudo-labeling. Methods such as LAPO \cite{sec2_b11} and Genie \cite{sec2_b12} follow a related pipeline while incorporating variational latent-action inference, which recovers latent actions from temporally adjacent observations, with evaluations conducted mainly in video-game domains that have restricted action spaces. DynaMo \cite{sec2_b13} and LAPA \cite{sec2_b14} train VAEs on unlabeled, in-domain robotic demonstrations, producing representations useful for downstream control. Notably, LAPA reports that direct projection from latent actions to real actions degrades performance, which we attribute to insufficient mutual information between the latent variables and the ground-truth actions. In this work, we adopt a variational information bottleneck formulation that increases this mutual information and, in turn, improves performance on downstream robotic tasks.

\section{METHOD}
\subsection{Variational Information Bottleneck for Latent Action Extraction}

We denote by $O_t\in\mathbb{R}^{C\times H\times W}$ the video frame at time $t$, and by $a_t\in\mathbb{R}^{D_a}$ the corresponding (unobserved) robot action.  Our goal is to learn a stochastic encoder
$$
  q_\phi(z|O_t)\;:\;\mathbb{R}^{C\times H\times W}\;\to\;\mathcal{N}\bigl(\mu_\phi(O_t),\,\Sigma_\phi(O_t)\bigr),
$$
that maps each frame $O_t$ to a latent action variable $z\in\mathbb{R}^{D_z}$.  We wish $z$ to satisfy two desiderata:
\begin{enumerate}
  \item \textbf{Predictivity:} $z$ should retain all information necessary to predict the next observation $O_{t+1}$.
  \item \textbf{Compression:} $z$ should discard irrelevant details of $O_t$, preserving only features that correlate with the true control $a_t$.
\end{enumerate}

These objectives can be cast in the Variational Information Bottleneck (VIB) framework \cite{sec3_b1}: we maximize the mutual information between $z$ and the future $O_{t+1}$ while penalizing the mutual information between $z$ and the input $O_t$.  Formally, we optimize
\begin{equation}
\begin{split}
\mathcal{L}_{\mathrm{VIB}}
&= -\,I\bigl(\mathcal{Z};O_{t+1}\bigr)
  + \beta\,I\bigl(Z;O_t\bigr),
\end{split}
\label{eq:vib_objective}
\end{equation}
where $\beta>0$ balances prediction accuracy against compression.

\subsection{Variational Lower Bounds.}\label{sec:variational_lower_bound}
Direct computation of the mutual informations in \eqref{eq:vib_objective} is intractable.  Introducing a Gaussian-likelihood decoder $p_\theta(X_{t+1}|z)$ and a prior $r(z)=\mathcal{N}(0,I)$, standard variational bounds give:
\begin{equation}
\begin{split}
I(Z;O_{t+1})
&\;\ge\;
\mathbb{E}_{O_t}\,
\mathbb{E}_{z\sim q_\phi(z| O_t)}
\bigl[\log p_\theta(O_{t+1}|z)\bigr] + \mathrm{const},
\end{split}
\label{eq:vib_pred}
\end{equation}
and
\begin{equation}
I(Z;O_t)
= \mathbb{E}_{O_t}\bigl[D_{\mathcal{KL}}\bigl(q_\phi(z|O_t)\,\|\,r(z)\bigr)\bigr].
\label{eq:vib_comp}
\end{equation}
Substituting \eqref{eq:vib_pred} and \eqref{eq:vib_comp} into \eqref{eq:vib_objective}, dropping constants, yields the tractable training loss:
\begin{equation}
\begin{split}
\mathcal{L}_{\mathrm{VIB}}
&= -\,\mathbb{E}_{O_t}\,
    \mathbb{E}_{z\sim q_\phi(z| O_t)}
      \bigl[\log p_\theta(O_{t+1}| z)\bigr] \\[-3pt]
&\quad + \beta\,
    \mathbb{E}_{O_t}\bigl[D_{\mathcal{KL}}\bigl(q_\phi(z| O_t)\,\|\,r(z)\bigr)\bigr].
\end{split}
\label{eq:vib_loss}
\end{equation}

\begin{remark}[\textbf{Implicit Action--Information Maximisation}]
Assume deterministic first-order dynamics $f$ and that the action
is \emph{recoverable} via a bijection $a_t=g(O_t,O_{t+1})$.
Then
\begin{equation}
  I(\mathcal Z;O_{t+1}| O_t)
  =I(\mathcal Z;a_t| O_t).
  \label{eq:cond_eq}
\end{equation}
The reconstruction term in the VIB loss is a variational \emph{lower}
bound on $I(\mathcal Z;O_{t+1}| O_t)$; minimising the loss therefore
maximises this conditional mutual information, and hence
$I(\mathcal Z;a_t| O_t).$
In other words, the latent code
$\mathcal Z$ learns to encode the true control signal
$a_t$ \emph{implicitly}, even though $a_t$ is never observed during
training.

\medskip
\noindent\textbf{From conditional to global mutual information.}
We next argue that the same objective \emph{approximately} increases the
global quantity $I(\mathcal Z;a_t)$ when the bottleneck coefficient
$\beta$ is sufficiently large.

For any three random variables the chain rule gives:
\begin{equation}
  I(\mathcal Z;a_t)
  =I(\mathcal Z;a_t| O_t)
   +I(\mathcal Z;O_t)
   -I(\mathcal Z;O_t| a_t).
  \label{eq:chain}
\end{equation}

The KL term in the VIB objective upper-bounds the mutual information between $\mathcal Z$ and $O_t$:
\begin{equation*}
  I(\mathcal Z;O_t)
  \le
  D_{\mathcal{KL}}\!\bigl(q_\phi(z| O_t)\,\|\,r(z)\bigr).
\end{equation*}
Choosing a large $\beta$ therefore forces
$$I(\mathcal Z;O_t)\!\to 0.$$

By assumption, each recorded action is a deterministic function of its
concurrent observation, $a_t=g(O_t,\cdot)$, adding $a_t$ cannot increase
information about $O_t$:
$$0\le I(\mathcal Z;O_t| a_t)\le I(\mathcal Z;O_t).$$
Hence $I(\mathcal Z;O_t)\!\to 0$ also drives
$I(\mathcal Z;O_t| a_t)\!\to 0$.

Substituting the two vanishing terms into~\eqref{eq:chain} yields
\begin{equation}
  I(\mathcal Z;a_t)
  \;\approx\;
  I(\mathcal Z;a_t| O_t)
  =I(\mathcal Z;O_{t+1}| O_t),
  \label{eq:approx}
\end{equation}
Thus, when $\beta$ is large, minimising the VIB loss not only maximises the conditional mutual information in \eqref{eq:cond_eq} but also promotes a \emph{globally} informative
latent code, making $\mathcal Z$ a faithful carrier of the underlying
control signal $a_t$.
\end{remark}

\subsection{Implementation Details}\label{sec:Implementation_details}

Our model comprises three modules: (i) an \textbf{encoder} $q_{\phi}(z| O_t)$ producing a Gaussian distribution over $z$, (ii) a \textbf{reparameterization} step sampling $z\sim q_{\phi}(z| O_t)$, and (iii) a \textbf{decoder} $p_{\theta}(O_{t+1}| z)$ that reconstructs the next observation (frame or flow).  All parameters are gathered in $(\phi,\theta)$.

\medskip\noindent\textbf{Encoder.}  
The encoder implements
\begin{equation*}
  q_{\phi}(z| O_t)
  = \mathcal{N}\bigl(\mu_{\phi}(O_t),\,\mathrm{diag}\,\sigma_{\phi}^2(O_t)\bigr),
\end{equation*}
where $\mu_{\phi}(\cdot)$ and $\sigma_{\phi}^2(\cdot)$ are the outputs of a learnable backbone followed by two small projection heads.

\medskip\noindent\textbf{Reparameterization.}  
We draw noise $\epsilon\sim\mathcal{N}(0,I)$ and form the latent
\begin{equation*}
  z \;=\; \mu_{\phi}(O_t)\;+\;\sigma_{\phi}(O_t)\,\epsilon.
\end{equation*}

\medskip\noindent\textbf{Decoder.}  
A learnable network $p_{\theta}$ maps each $z\in\mathbb{R}^{D_z}$ back to the observation space:
\begin{equation*}
  \hat O_{t+1} = p_{\theta}(z).
\end{equation*}

\medskip\noindent\textbf{Training Loss.}  
We minimize the VIB objective in (\ref{eq:vib_loss}) with 
\begin{equation*}
  -\log p_{\theta}(O_{t+1}| z)
  \approx \bigl\|\,\hat O_{t+1} - O_{t+1}\bigr\|_{2}^{2},
  \quad
  r(z)=\mathcal{N}(0,I).
\end{equation*}

\medskip
\begin{algorithm}[t]
\caption{VIB Training for Latent–Action Extraction}
\label{alg:vib}
\begin{algorithmic}[1]
\State {\bfseries Input:}\quad Sequence of frames $\{O_t,O_{t+1}\}_{t=1}^N$  
\State {\bfseries Hyperparameters:}\quad $\beta>0$, learning rate $\eta$
\State Initialize encoder–decoder parameters $(\phi,\theta)$
\For{each epoch}
  \For{each minibatch $\{O_t,O_{t+1}\}$}
    \State \# 1. Encode current frame
    \State $h \gets \mathrm{Backbone}_\phi(O_t)$
    \State $\mu,\log\sigma^2\gets \mathrm{Heads}_\phi(h)$
    \State \# 2. Sample latent via reparameterization
    \State $\epsilon\sim\mathcal{N}(0,I)$ 
    \State $z\gets \mu + \exp(\tfrac12\log\sigma^2)\,\epsilon$
    \State \# 3. Decode to predict next frame
    \State $\hat O_{t+1}\gets \mathrm{Decoder}_\theta(z)$
    \State \# 4. Compute losses
    \State $\mathcal L_{\mathrm{rec}} \gets \|O_{t+1}-\hat O_{t+1}\|_2^2$
    \State $\mathcal L_{\mathrm{KL}}  \gets \mathcal{KL}\bigl(\mathcal{N}(\mu,\sigma^2)\|\mathcal{N}(0,I)\bigr)$
    \State $\mathcal L \gets \mathcal L_{\mathrm{rec}} + \beta\,\mathcal L_{\mathrm{KL}}$
    \State \# 5. Gradient descent
    \State $(\phi,\theta)\gets (\phi,\theta) \;-\;\eta\,\nabla_{\phi,\theta}\,\mathcal L$
  \EndFor
\EndFor
\State {\bfseries Return:} Trained encoder $q_\phi(z| O)$
\end{algorithmic}
\end{algorithm}

\begin{remark}[\textbf{Model Simplicity and Novelty}]
Although our approach builds on the classical variational autoencoder paradigm, it is distinguished by a principled Variational Information Bottleneck formulation tailored to latent–action extraction.  In contrast to LAPA, which derives latent actions from the difference of two frame‐level feature embeddings, we instead learn a single‐frame encoder whose bottleneck representation is explicitly optimized to reconstruct the subsequent frame (or its optical flow).  This simple yet theoretically grounded shift not only adheres to the VIB objective developed above, but—as we demonstrate in Section \ref{sec:experiments}—yields substantially higher mutual information with the true control signals, resulting in more informative and predictive latent actions.
\end{remark}

\subsection{Latent–to–Action Mapping}

While the VIB framework extracts a continuous latent \(z\) that captures task‐relevant control information, we still require a mechanism to convert \(z\) into the true robot action \(a\in\mathbb{R}^{D_a}\). In this work, we follow two main paradigms:

\begin{enumerate}
  \item \textbf{Direct projection (LAPO‐style \cite{sec2_b11}).}  An action head \(g_\psi^{\text{LAPO}}: \mathbb{R}^{D_z}\to\mathbb{R}^{D_a}\) is trained so that:
  \[
    a_t = g_\psi^{\text{LAPO}}\bigl(z_t\bigr),\quad z_t\sim q_\phi(z| O_t, O_{t+1}).
  \]
  
   \item \textbf{Discrete‐index decoding (LAPA‐style \cite{sec2_b14}).} An encoder–decoder pair is trained to predict the quantized index:
   \[
        h_t = f_{\mathrm{enc}}^{\mathrm{LAPA}}(O_t), 
   \quad 
   p_t = f_{\mathrm{dec}}^{\mathrm{LAPA}}(h_t), 
   \]
   where $p_{t,i} = P\bigl(i_t=i | O_t,O_{t+1}\bigr)$ and \(i_t=\mathrm{Index}\bigl(q_\phi(z | O_t,O_{t+1})\bigr)\). After training,  \(f^{\mathrm{LAPA}}_{\mathrm{dec}}\) is discard.  A new action head \(h_\psi\) is then learned to map the frozen features from encoder \(f^{\mathrm{LAPA}}_{\mathrm{enc}}\) directly to the discretized action. 

\end{enumerate}

In our implementation, we adopt the LAPO‐style by projecting the latent variable $z$ directly into the action $a$. For the LAPA‐style, we simply freeze the encoder introduced in Sec. \ref{sec:Implementation_details} and use it as $ f_{\mathrm{enc}}^{\mathrm{LAPA}}(O_t)$.

\section{EXPERIMENTS}

\subsection{Numerical Analysis}\label{sec:experiment_numerical}

In this section, we empirically evaluate the ability of our VIB‐based latent–action extractor to capture true action information from purely unlabelled video demonstrations.  We conduct experiments on the eight MetaWorld manipulation tasks, each with 100 expert demos per task.  We measure two complementary metrics between the learned latent codes $\mathcal{Z}$ and the ground‐truth robot commands $\mathcal{A}$:
\begin{itemize}
  \item \textbf{Pearson correlation:} the linear correlation coefficient $r(z_i,a_j)$ for each latent dimension $z_i$ and action channel $a_j$.
  \item \textbf{Mutual information capture ratio:} the discrete mutual information $I(z_i; a_j)$ normalized by the marginal entropy $H(a_j)$.
\end{itemize}

We study two factors that influence the amount of action information extracted:
\begin{enumerate}
  \item \emph{Bottleneck strength.}  How does varying the VIB penalty weight $\beta$ affect the trade‐off between compression and predictive power?
  \item \emph{Temporal gap.}  How does the frame‐pair offset $k$ (i.e.\ comparing frames at time $t$ and $t+k$) influence the quality of the extracted latent actions?
\end{enumerate}

\begin{table}[t]\scriptsize
  \centering
  \caption{Maximum $|r(z_i,a_j)|$ and MI capture ratio $I(z_i;a_j)/H(a_j)$ per action channel for our VIB‐based extractor on MetaWorld.}
  \label{tab:alignment_metrics}
  \begin{tabular}{lcccc}
    \toprule
    Metric & dim~0  & dim~1  & dim~2  & dim~3 \\
    \midrule
    $\max |r(z_i,a_j)|$           & 0.562 & 0.776 & 0.565 & 0.762 \\
    $\max I(z_i;a_j)/H(a_j)$ (\%) & 51.5  & 56.1  & 53.7  & 80.2  \\
    \bottomrule
  \end{tabular}
\end{table}

\begin{figure}[t]
  \centering
  \includegraphics[width=\linewidth]{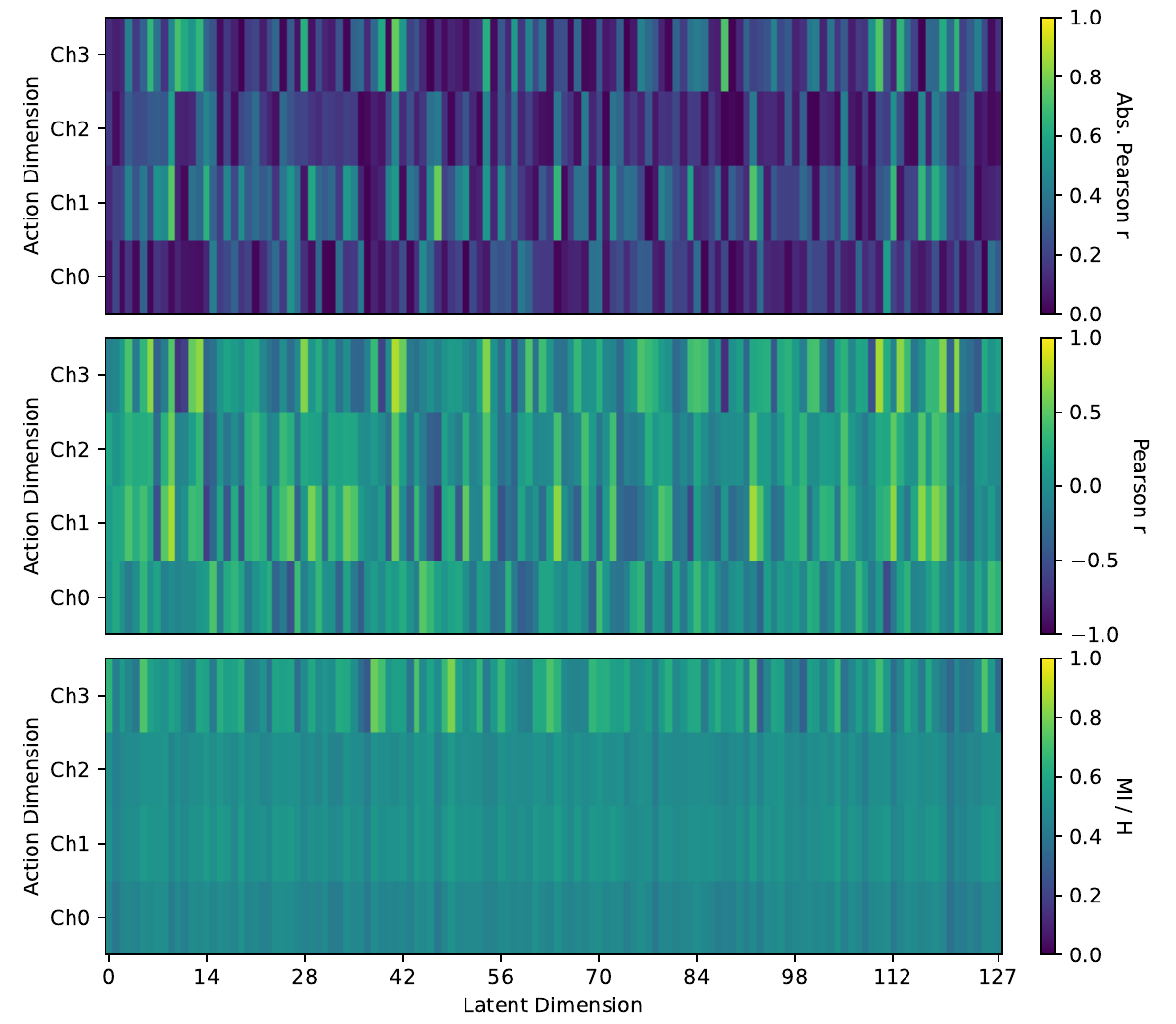}
  \caption{Heatmaps of latent–action alignment for our VIB‐based extractor on MetaWorld.  
    \textbf{Top:} absolute Pearson correlation $|r(z_i,a_j)|$.  
    \textbf{Bottom:} normalized MI capture ratio $I(z_i; a_j)/H(a_j)$.  
    Compared to the LAPA baseline, our method exhibits uniformly higher correlations and capture ratios across all latent–action pairs.}
  \label{fig:alignment_heatmap}
\end{figure}

\begin{figure}[t]
  \centering
  \includegraphics[width=0.95\linewidth]{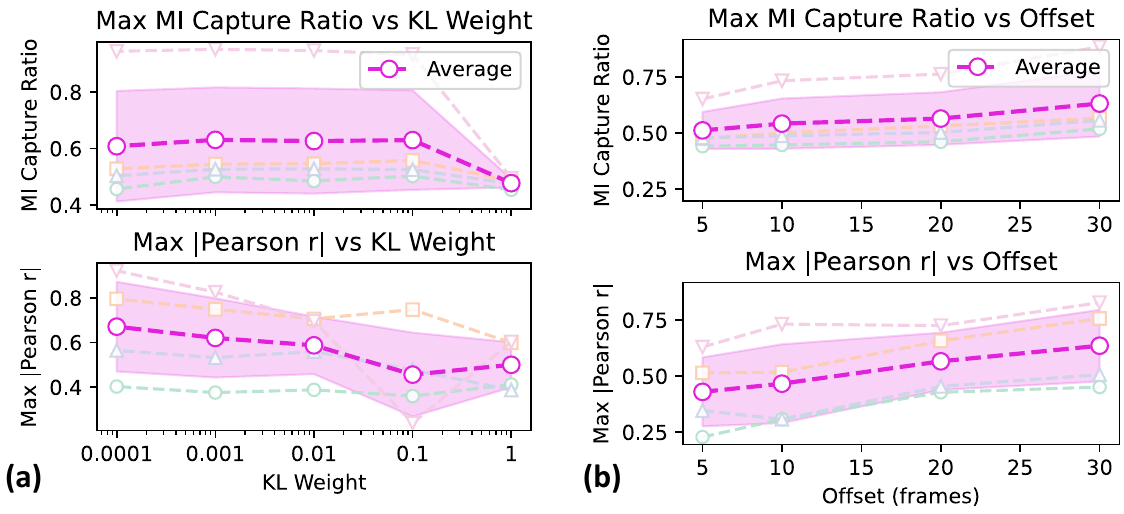}
  \caption{Effect of KL weight (\(\beta\)) and frame offset on latent--action alignment. 
(a) Top: maximum MI capture ratio vs. \(\beta\); bottom: maximum absolute Pearson correlation \(|r|\) vs. \(\beta\). 
(b) Top: maximum MI capture ratio vs. frame offset; bottom: maximum absolute Pearson correlation \(|r|\) vs. frame offset. 
Dashed colored lines denote individual action dimensions, the solid purple line indicates the mean across dimensions, and the shaded band represents \(\pm 1\) standard deviation.}
  \label{fig:numerical}
\end{figure}

In Fig. \ref{fig:numerical}, we observe that as the KL penalty weight $\beta$ increases, the maximum MI capture ratio first increase slowly and decrease when $\beta$ is too big. This result is aligned with our remark in Sec. \ref{sec:variational_lower_bound}. For larger $\beta$, the gap beween the conditional mutual information and the global mutual information become smaller. However, overly large $\beta$ (e.g. $\beta=1.0$)  values force the encoder to discard not only nuisance variation but also useful control signals, leading to a decreasement in  the maximum MI capture ratio. However, we found that the maximum Pearson correlation $|r|$ decreases steadily as the KL penalty weight
$\beta$ increases. This trend indicates that a stronger information bottleneck, while encouraging more compact latent representations, also restricts the capacity of $\mathcal{Z}$ to retain a linear relationship to the action information.

Conversely, in Fig. \ref{fig:numerical} the metrics both rise as the frame‐pair offset $k$ grows.  Larger temporal gaps amplify the magnitude of state changes between frames, making motion dynamics more pronounced in the input and thus easier for the latent extractor to capture.  This result suggests that choosing an appropriate offset can enhance the signal‐to‐noise ratio for action inference.

In Table \ref{tab:alignment_metrics}, we compare the per‐channel maximum absolute Pearson correlation and mutual‐information capture ratio achieved by our VIB‐based extractor against the LAPA baseline on MetaWorld.  Our method attains $|r|$ values of 0.56, 0.78, 0.57, and 0.76 (vs.\ 0.40, 0.48, 0.57, and 0.50 for LAPA) and MI capture ratios of 51.5\%, 56.1\%, 53.7\%, and 80.2\% (vs.\ 13.2\%, 18.2\%, 16.6\%, and 18.4\%).  Fig. \ref{fig:alignment_heatmap} visualizes these gains as heatmaps, showing uniformly stronger alignment across all latent–action pairs.

\subsection{Few-shot policy learning}

\begin{figure*}[!t] 
  \centering
  \includegraphics[width=0.95\textwidth]{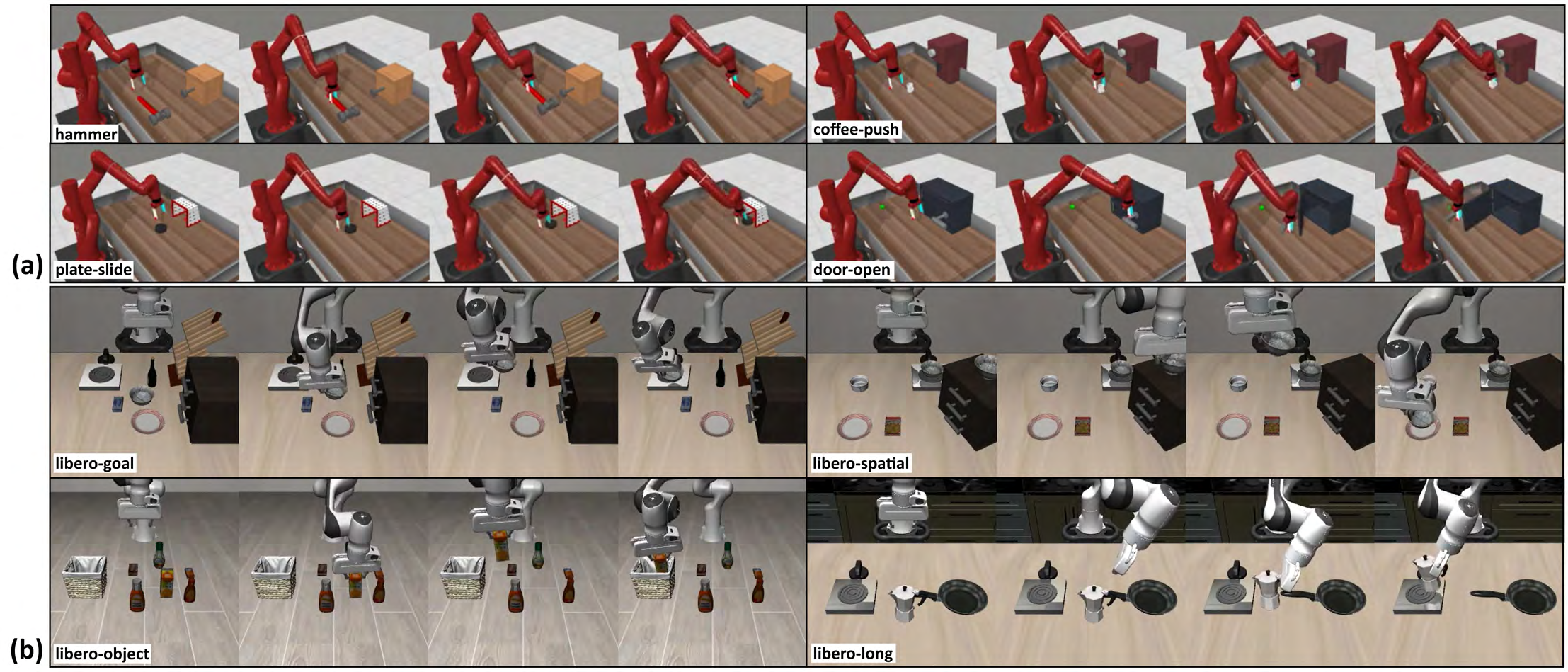}
  \caption{%
    Snapshots of different Metaworld tasks.
  }
  \label{fig:metaworld-result}
\end{figure*}

In this section, we evaluate the few‐shot learning performance on the \textbf{MetaWorld} \cite{sec1_metaworld} and \textbf{Libero} \cite{sec3_libero} benchmarks. For MetaWorld, we select eight representative tasks including \emph{assembly, button-press, coffee-push, door-open, hammer, pick-place, plate-slide, push}. For each task, we collect 50 demonstration episodes and use all video frames (without action labels) to train the latent‐action model described in Sec. \ref{sec:Implementation_details}. For Libero, we train the same model on the full set of demonstrations provided by the official dataset. 

Our encoder follows the Vision Transformer (ViT) framework: we use ViT‑Tiny for MetaWorld and ViT‑Base for Libero. The decoder is a three‐layer convolutional network with channel widths $[64,\,32,\,3]$. All models are optimized with Adam at a learning rate of $1\times10^{-4}$. To incorporate semantic task information in Libero, we encode each task description using the T5‑Base model \cite{sec3_t5}, and condition both the latent prediction and reconstruction on the resulting text embeddings. All latent action models are trained for 30,000 epochs. For few‐shot learning on MetaWorld, we use a two‐layer residual‐block MLP with hidden size 512 as the action head. For Libero, we adopt the diffusion‐policy implementation in \cite{sec3_diffusion_policy}. The action head is trained for 300k epochs on MetaWorld and 2M epochs on Libero.

To validate our approach, we compare against several state‐of‐the‐art baselines on both benchmarks. On MetaWorld, we evaluate \textbf{ResNet\-T} \cite{sec3_libero}, \textbf{Flow-Matching (FM)} \cite{sec3_flow_matching}, \textbf{LAPO} \cite{sec2_b11} and \textbf{LAPA} \cite{sec2_b14} to demonstrate the effectiveness of our method. On Libero, we further assess \textbf{Diffusion Policy} \cite{sec3_diffusion_policy}, \textbf{Octo} \cite{sec3_octo}, \textbf{OpenVLA} \cite{sec3_openvla} to showcase its efficiency.

\begin{figure*}[!t] 
  \centering
  \includegraphics[width=0.95\textwidth]{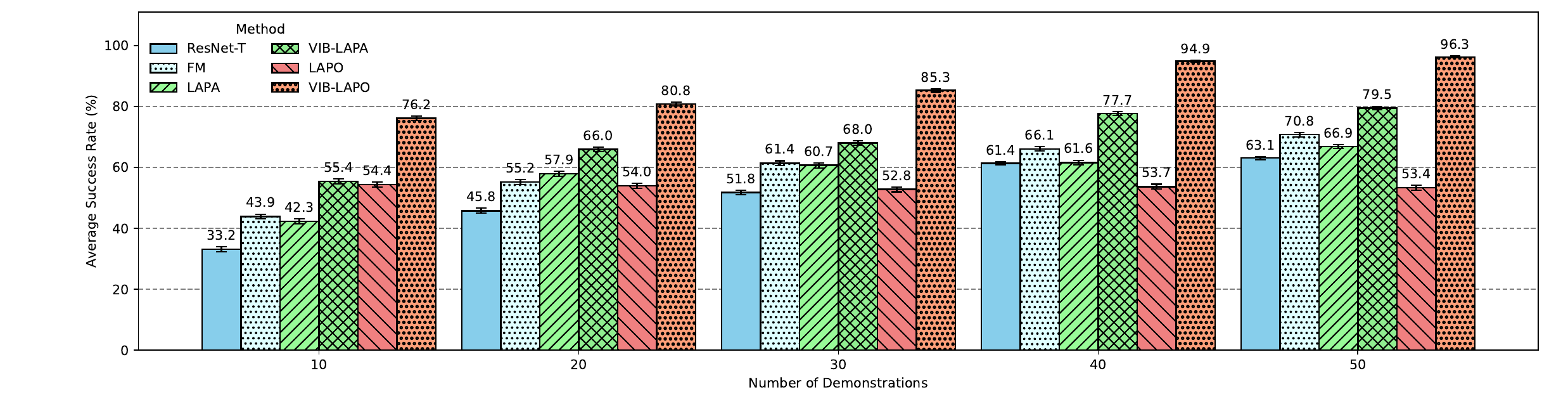}
  \caption{%
    Performance of VIB-LAPA and VIB-LAPO compared to original LAPA, LAPO, and from-scratch baselines across different numbers of demonstration trajectories.
  }
  \label{fig:metaworld-result}
\end{figure*}

As shown in Fig. \ref{fig:metaworld-result}, when the number of demonstration trajectories is small, pretrained methods such as LAPA that leverage unlabeled videos substantially improve data efficiency compared to training from scratch. However, as the number of demonstrations increases, the performance of LAPA (and LAPO) plateaus and eventually falls behind the from-scratch baselines, due to their latent variables capturing relatively low mutual information with the true actions. In contrast, our VIB-LAPA and VIB-LAPO models—which explicitly maximize the mutual information between latent representations and control signals—consistently outperform both the original LAPA/LAPO and the from-scratch methods across all demonstration regimes. This result confirms the effectiveness of unsupervised pretraining on unlabeled video data and the critical role of high latent-action mutual information.

The Libero result is shown in Table \ref{tab:libero}. The result shows that our method  attains an average success rate on par with OpenVLA~\cite{sec3_openvla} while using only {4.4\%} (306M vs 7b) of OpenVLA’s parameters, demonstrating strong parameter efficiency.
Moreover, compared with LAPA~\cite{sec2_b14}, our approach improves performance across all suites as well as the overall average, supporting the effectiveness of extracting latent actions that preserve higher mutual information with the ground‐truth actions.


\begin{table}[t]
  \centering
  \caption{Results on Libero task suites (success rate, \%). ${*}$LAPA is reproduced using the Prismatic-7B VLM \cite{sec3_univla}.}
  \label{tab:libero}
  \footnotesize
  \setlength{\tabcolsep}{4pt}
  \resizebox{\columnwidth}{!}{%
  \begin{tabular}{l|ccccc} 
    \toprule
    \textbf{Method}  & \textbf{Spatial} & \textbf{Object} & \textbf{Goal} & \textbf{Long} & \textbf{Average} \\
    \midrule
    Diffusion Policy \cite{sec3_diffusion_policy} & 78.3 & \textbf{92.5} & 68.3 & 50.5 & 72.4 \\
    Octo \cite{sec3_octo}             & 78.9 & 85.7 & 84.6 & 51.1 & 75.1 \\
    OpenVLA \cite{sec3_openvla}          & \textbf{84.7} & 88.4 & 79.2 & 53.7 & 76.5 \\
    \rowcolor{red!10}
    LAPA$^{*}$ \cite{sec2_b14}       & 73.8 & 74.6 & 58.8 & 55.4 & 65.7 \\
    \rowcolor{green!10}
    Ours         & 75.4 & 87.4 & \textbf{87.8} & \textbf{56.7} & \textbf{76.9} \\
    \bottomrule
  \end{tabular}%
  }
\end{table}




\section*{ACKNOWLEDGMENT}


\end{document}